\pdfoutput=1

\documentclass[11pt]{article}

\usepackage{acl}

\usepackage{times}
\usepackage{latexsym}
\usepackage{colortbl}
\usepackage[T1]{fontenc}

\usepackage[utf8]{inputenc}

\usepackage{microtype}
\usepackage{booktabs}
\usepackage{graphicx}
\usepackage{amsmath}
\usepackage[noabbrev,capitalize,nameinlink]{cleveref}
\usepackage[ruled,vlined,linesnumbered]{algorithm2e}
\usepackage{tabularx}
\usepackage{tikz}
\usepackage[draft]{minted}
\usepackage{float}
\usepackage{listings}
\usepackage{fontawesome5}
\newcommand*\circled[1]{\tikz[baseline=(char.base)]{\node[shape=circle,draw,inner sep=.6pt] (char) {#1};}}
\newcommand{\std}[2]{{#1}{\footnotesize$\pm${#2}}}
\newcommand{\bertb}{BERT$_\text{base}$}
\newcommand{\bertl}{BERT$_\text{large}$}

\newcommand{\bartl}{BART$_\text{large}$}

\lstdefinestyle{json}{
  basicstyle=\ttfamily,
  showstringspaces=false,
  breaklines=true,
  frame=single,
  backgroundcolor=\color{gray!10},
  language=Python, 
  morekeywords={true, false, null},
}

\title{Entity Linking in the Job Market Domain}

\author{Mike Zhang\textsuperscript{\faCompass}\textsuperscript{\faRobot} \hspace{1em}
Rob van der Goot\textsuperscript{\faCompass}\textsuperscript{\faRobot} \hspace{1em}
Barbara Plank\textsuperscript{\faCompass}\textsuperscript{\faMountain}\textsuperscript{\faHiking}\\
\textsuperscript{\faCompass}Department of Computer Science, IT University of Copenhagen, Denmark\\
\textsuperscript{\faRobot}Pioneer Centre for Artificial Intelligence, Copenhagen, Denmark \\
\textsuperscript{\faMountain}MaiNLP, Center for Information and Language Processing, LMU Munich, Germany \\
\textsuperscript{\faHiking}Munich Center for Machine Learning (MCML), Munich, Germany \\
{\tt mikejj.zhang@gmail.com }}

\begin{document}
\maketitle
\begin{abstract}
In Natural Language Processing, entity linking (EL) has centered around Wikipedia, but remains underexplored for the job market domain.
Disambiguating skill mentions can help us to get insight into the labor market demands. In this work, we are the first to explore EL in this domain, specifically targeting the linkage of occupational skills to the ESCO taxonomy~\cite{le2014esco}. Previous efforts linked coarse-grained (full) sentences to a corresponding ESCO skill. In this work, we link more fine-grained span-level mentions of skills. We tune two high-performing neural EL models, a bi-encoder~\cite{wu-etal-2020-scalable} and an autoregressive model~\cite{cao2021autoregressive}, on a synthetically generated mention--skill pair dataset and evaluate them on a human-annotated skill-linking benchmark. Our findings reveal that both models are capable of linking implicit mentions of skills to their correct taxonomy counterparts. Empirically, BLINK outperforms GENRE in strict evaluation, but GENRE performs better in loose evaluation (accuracy@$k$).\footnote{The source code and data can be found at \url{https://github.com/mainlp/el_esco}}

\end{abstract}

\section{Introduction}
Labor market dynamics, influenced by technological changes, migration, and digitization, have led to the availability of job descriptions (JD) on platforms to attract qualified candidates~\citep{brynjolfsson2011race,brynjolfsson2014second,balog2012expertise}. 
It is important to extract and link surface form skills to a unique taxonomy entry, allowing us to quantify the current labor market dynamics and determine the demands and needs. We attempt to tackle the problem of \emph{entity linking} (EL) in the job market domain, specifically the linking of fine-grained span-level skill mentions to a specific taxonomy entry. 

Generally, EL is the task of linking mentions of entities in unstructured text documents to their respective unique entities in a knowledge base (KB), most commonly Wikipedia~\citep{he-etal-2013-learning}. Recent models address this problem by producing entity representations from a (sub)set of KB information, e.g., entity descriptions~\citep{logeswaran-etal-2019-zero,wu-etal-2020-scalable}, fine-grained entity types~\citep{raiman2018deeptype,onoe2020fine,ayoola-etal-2022-refined}, or generation of the input text autoregressively~\citep{cao2021autoregressive,de2022multilingual}.

For skill linking specifically, we use the European Skills, Competences, Qualifications and Occupations (ESCO;~\citealp{le2014esco}) taxonomy due to its comprehensiveness. Previous work classified spans to its taxonomy code via multi-class classification~\citep{zhang-etal-2022-kompetencer} without surrounding context and neither the full breadth of ESCO.~\citet{gnehm-etal-2022-fine} approaches it as a sequence labeling task, but only uses more coarse-grained ESCO concepts, and not the full taxonomy. Last, others attempt to match the full sentence to their respective taxonomy title~\citep{decorte2022design,decorte2023extreme,clavie2023large}. 

The latter comes with a limitation: The taxonomy title does not indicate which subspan in the sentence it points to, without an exact match. We define this as an \emph{implicit} skill, where mentions (spans) in the sentence do not have an exact string match with a skill in the ESCO taxonomy. The differences can range from single tokens to entire phrases. For example, we can link ``being able to work together'' to ``plan teamwork''.\footnote{See example here: \url{https://t.ly/3VUJG}.} If we know the exact span, this implicit skill can be added to the taxonomy as an alternative choice for the surface skill. As a result, this gives us a more nuanced view of the labor market skill demands. Therefore, we attempt to train models to the linking of both implicit and explicit skill mentions.

\paragraph{Contributions.} Our findings can be summarized as follows: \circled{1} We pose the task of skill linking as an entity linking problem, showing promising results of linking with two entity linking systems.
\circled{2} We present a qualitative analysis showing that the model successfully links implicit skills to their respective skill entry in ESCO.

\begin{table}[t]
    \centering
    \resizebox{.85\linewidth}{!}{
    \begin{tabular}{lrrr}
    \toprule
             & Instances & Unique Titles & \texttt{UNK}\\
    \midrule
    Train    & 123,619   & 12,984   & 14,641\\
    Dev.     & 480       & 149      & 233 \\
    Test     & 1,824     & 455      & 813\\
    \bottomrule
    \end{tabular}}
    \caption{\textbf{Data Statistics.} Data distribution of train, dev, and test splits. \texttt{UNK} indicates skills mentions that are not linked to a corresponding taxonomy title.
    }
    \label{tab:data}
\end{table}

\section{Methodology}

\paragraph{Definition.}
In EL, we process the input document $\mathcal{D} = \{w_1, \ldots, w_r\}$, a collection of entity mentions denoted as $\mathcal{MD} = \{m_1, \ldots, m_n\}$, and a KB, ESCO in our case: $\mathcal{E} = \{e_1, \ldots, e_{13890}, \text{\texttt{UNK}}\}$. The objective of an EL model is to generate a list of mention-entity pairs $\{(m_i, e_i)\}_{i=1}^{n}$, where each entity $e$ corresponds to an entry in a KB.
We assume that both the titles and descriptions of the entities are available, which is a common scenario in EL research~\cite{ganea-hofmann-2017-deep,logeswaran-etal-2019-zero,wu-etal-2020-scalable}. We also assume that each mention in the document has a corresponding valid gold entity present in the knowledge base, including \texttt{UNK}. This scenario is typically referred to as ``in-KB evaluation''. Similar to prior research efforts~\cite{logeswaran-etal-2019-zero,wu-etal-2020-scalable}, we also presuppose that the mentions within the document have already been tagged. 

\paragraph{Data.}
\looseness=-1
We use ESCO titles as ground truth labels, containing 13,890 skills.\footnote{Per version 1.1.1, accessed on 01 August 2023.}
\cref{tab:data} presents the train, dev, and test data in our experiments. We leverage the train set introduced by~\citet{decorte2023extreme}\footnote{\url{https://t.ly/edqkp}} along with the dev and test sets provided in~\citet{decorte2022design}.\footnote{\url{https://t.ly/LcqQ7}} 
The train set is synthetically generated by~\citet{decorte2023extreme} with the \texttt{gpt-3.5-turbo-0301} model~\cite{openai-chat}. Specifically, this involves taking each skill from ESCO and prompting the model to generate sentences resembling JD sentences that require that particular skill. The dev and test splits, conversely, are derived from actual job advertisements sourced from the study by~\citet{zhang-etal-2022-skillspan}. These JDs are annotated with spans corresponding to specific skills, and these spans have subsequently been manually linked to ESCO, as described in the work of~\citet{decorte2022design}. In cases where skills cannot be linked, two labels are used, namely \texttt{UNDERSPECIFIED} and \texttt{LABEL NOT PRESENT}. For the sake of uniformity, we map both of these labels to a generic \texttt{UNK} tag. We used several heuristics based on Levenshtein distance and sentence similarity to find the exact subspans if it exceeds certain thresholds, otherwise, it is \texttt{UNK}. This process is outlined in~\cref{app:preprocess}. In addition, some data examples can be found in~\cref{app:examples}. The number of \texttt{UNK}s in the data is also in~\cref{tab:data}. During inference, the \texttt{UNK} title is a prediction option for the models. 

\begin{table*}[t]
    \centering
    \resizebox{\textwidth}{!}{%
    \begin{tabular}{lrrrrrrr}
    \toprule
         & \textbf{Train Source} & \textbf{Acc@1}  & \textbf{Acc@4} & \textbf{Acc@8} & \textbf{Acc@16} & \textbf{Acc@32} \\
    \midrule
Random                      & & \std{0.22}{0.00} & \std{0.88}{0.00} & \std{1.76}{0.00} & \std{3.52}{0.00} & \std{7.04}{0.00} \\
TF-IDF                      & & \std{2.25}{0.00} & \cellcolor{lightgray} & \cellcolor{lightgray} & \cellcolor{lightgray} & \cellcolor{lightgray}\\
\midrule
BLINK (\texttt{bert-base})   &  ESCO             & \std{12.74}{0.49} & \std{22.81}{0.79} & \std{27.70}{0.82} & \std{32.44}{1.33} & \std{36.46}{1.07}\\
BLINK (\texttt{bert-large})  &  ESCO             & \std{12.77}{0.94} & \std{22.58}{1.47} & \std{27.24}{1.23} & \std{31.75}{0.89} & \std{36.10}{1.28}\\
BLINK (\texttt{bert-large})  &  Wiki (0-shot)    & \std{23.30}{0.00} & \textbf{\std{32.89}{0.00}} & \textbf{\std{38.16}{0.00}} & \std{42.60}{0.00} & \std{45.56}{0.00}\\
BLINK (\texttt{bert-large})  &  Wiki + ESCO      & \textbf{\std{23.55}{0.14}} & \std{32.63}{0.16} & \std{37.38}{0.09} & \textbf{\std{43.25}{0.13}} & \textbf{\std{48.98}{0.21}} \\\midrule
GENRE (\texttt{bart-base})           &  ESCO             & \std{1.47}{0.05} & \std{4.84}{1.74} & \std{10.46}{6.81} & \std{11.30}{4.18} & \std{15.51}{4.62}\\
GENRE (\texttt{bart-large})          &  ESCO             & \std{2.33}{0.44} & \std{5.74}{1.43} & \std{8.18}{2.21} & \std{11.13}{2.42} & \std{15.26}{2.66}\\
GENRE (\texttt{bart-large})          &  Wiki (0-shot)    & \std{6.91}{0.00} & \std{12.34}{0.00} & \std{15.52}{0.00} & \std{21.60}{0.00} & \std{33.17}{0.00}\\
GENRE (\texttt{bart-large})          &  Wiki + ESCO      & \textbf{\std{11.48}{0.41}} & \textbf{\std{21.26}{0.43}} & \textbf{\std{27.40}{0.78}} & \textbf{\std{37.21}{0.69}}  & \textbf{\std{49.78}{1.05}} \\
\bottomrule
    \end{tabular}%
    }    
    \caption{\textbf{Skill Linking Results.} We show the results of the various models used. There are two \texttt{base} and four \texttt{large} models. Training sources are either ESCO or a combination of Wikipedia and ESCO. The results are the average and standard deviation over five seeds. For the 0-shot setup, we apply the fine-tuned models from the work of~\citet{wu-etal-2020-scalable} and~\citet{cao2021autoregressive} to the ESCO test set once. We have a random and a TF-IDF-based baseline.}
    \label{tab:results}
\end{table*}

\paragraph{Models.}
We use two EL models, selected for their robust performance in EL on Wikipedia.
\footnote{For the hyperparameter setups, we refer to~\cref{app:hyperparams}.}

\paragraph{BLINK~\cite{wu-etal-2020-scalable}.}
BLINK uses a bi-encoder architecture based on BERT~\cite{devlin-etal-2019-bert}, for modeling pairs of mentions and entities. The model processes two inputs:
\begin{align*}
&\text{\texttt{[CLS]}} \hspace{0.5mm} \text{ctxt}_{\text{l}} \hspace{0.5mm}\text{\texttt{[S]}} \hspace{0.5mm} \text{mention} \hspace{0.5mm} \text{\texttt{[E]}} \hspace{0.5mm} \text{ctxt}_{\text{r}}  \hspace{0.5mm}\text{\texttt{[SEP]}}
\end{align*}
Where ``mention'', ``ctxt$_\text{l}$'', and ``ctxt$_\text{r}$'' corresponds to the wordpiece tokens of the mention, the left context, and the right context. The mention is denoted by special tokens \texttt{[S]} and \texttt{[E]}. The entity and its description are structured as follows:
\begin{align*}
&\text{\texttt{[CLS]}} \hspace{0.5mm} \text{title} \hspace{0.5mm} \text{\texttt{[ENT]}} \hspace{0.5mm} \text{description} \hspace{0.5mm}\text{\texttt{[SEP]}}
\end{align*}
Here, ``title'' and ``description'' represent the wordpiece tokens of the skills' title and description, respectively. \texttt{[ENT]} is a special token to separate the two representations. We train the model to maximize the dot product of the \texttt{[CLS]} representation of the two inputs, for the correct skill in comparison to skills within the same batch. For each training pair $(m_i, e_i)$, the loss is computed as
$\mathcal{L}\left(m_i, e_i\right) = -\operatorname{s}\left(m_i, e_i\right) + \log \sum_{j=1}^B \exp \left(\operatorname{s}\left(m_i, e_j\right)\right)$,
where the objective is to minimize the distance between $m_i$ and $e_i$ while encouraging the model to assign a higher score to the correct pair and lower scores to randomly sampled incorrect pairs. Hard negatives are also used during training, these are obtained by finding the top 10 predicted skills for each training example. These extra hard negatives are added to the random in-batch negatives.

\paragraph{GENRE~\cite{cao2021autoregressive}.} GENRE formulates EL as a retrieval problem using a sequence-to-sequence model based on BART~\cite{lewis-etal-2020-bart}. This model generates textual entity identifiers (i.e., skill titles) and ranks each entity $e \in \mathcal{E}$ using an autoregressive approach:
$\operatorname{s}(e \mid x) = p_\theta(y \mid x)=\prod_{i=1}^N p_\theta\left(y_i \mid y_{<i}, x\right)$,
where $y$ represents the set of $N$ tokens in the identifier of entity $e$ (i.e., entity tile), and $\theta$ denotes the model parameters. During decoding, the model uses a constrained beam search to ensure the generation of valid identifiers (i.e., only producing valid titles that exist within the KB, including \texttt{UNK}). 

\paragraph{Setup.} We train a total of six models: for BLINK, these are \bertb{} and \bertl{} (uncased;~\citealp{devlin-etal-2019-bert}) trained on ESCO, and another large version trained on Wikipedia and ESCO sequentially.
GENRE has the same setup, but then with BART~\cite{lewis-etal-2020-bart}. Additionally, we apply the released models from both BLINK and GENRE (large, trained on Wikipedia) in a zero-shot manner and evaluate their performance. The reason we use Wikipedia-based models is that we hypothesize this is due to many skills in ESCO also having corresponding Wikipedia pages (e.g., Python\footnote{\url{https://en.wikipedia.org/wiki/Python_(programming_language)}} or teamwork\footnote{\url{https://en.wikipedia.org/wiki/Teamwork}}), thus could potentially help linking. Next, to address unknown entities (\texttt{UNK}), we include them as possible label outputs. 

For evaluation, we assess the accuracy of generated mention-entity pairs in comparison to the ground truth. Here, we use the evaluation metric Accuracy@$k$, following prior research~\citep{logeswaran-etal-2019-zero,wu-etal-2020-scalable, zaporojets2022tempel}. We calculate the correctness between mentions and entities in the KB as the sum of correct hits or true positives (TP) if the ground truth for instance $i$ is in the top-$k$ predictions, formally:

\begin{equation}
\begin{aligned}
\text{Accuracy@$k$} = \frac{1}{n} \sum_{i=1}^{n} \text{TP in top-$k$ for instance $i$.}
\end{aligned}
\end{equation}

\newcolumntype{b}{X}
\newcolumntype{s}{>{\hsize=.6\hsize}X}
\begin{table*}[t]
    \centering
    \small
    \resizebox{\linewidth}{!}{
    \begin{tabularx}{\linewidth}{bss}
    \toprule
    \textbf{Mention}                                                                                                                       &    \textbf{BLINK}                                                        &   \textbf{GENRE}                                    \\
    \midrule
    \circled{1} Work in a way that is \textcolor{purple}{\textbf{patient-centred}} and inclusive.                                             &    \textcolor{teal}{person centred care} (K0913)                    & \textcolor{red}{work in an organised manner} (T) \\    \midrule
    \circled{2}    You can \textcolor{purple}{\textbf{ride a bike}}.                                                                             &    \textcolor{red}{sell bicycles} (S1.6.1)                      & \textcolor{teal}{drive two-wheeled vehicles} (S8.2.2)              \\    \midrule
    \circled{3} It is expected that you are a super user of the \textcolor{purple}{\textbf{MS office tools}}.                                 &    \textcolor{teal}{use Microsoft Office} (S5.6.1)                       & \textcolor{red}{tools for software configuration management} (0613) \\    \midrule
    \circled{4} \textcolor{purple}{\textbf{Picking and packing}}.                                                                             &    \textcolor{red}{carry out specialised packing for customers} (S6.1.3) & \textcolor{teal}{perform loading and unloading operations} (S6.2.1)\\    \midrule
    \circled{5} You are expected to be able to further \textcolor{purple}{\textbf{develop your team}} - both personally and professionally. \textcolor{orange}{GOLD: \textbf{manage a team} (S4.8.1)}  &    \textcolor{red}{manage personal professional development} (S1.14.1)   & \textcolor{red}{shape organisational teams based on competencies} (S4.6.0) \\ \midrule
    \circled{6} Our games are developed using Unity so we expect all our programmers to have solid knowledge of mobile game development in Unity3D and \textcolor{purple}{\textbf{C\#}}.  & \textcolor{teal}{C\#} (K0613)        & \textcolor{teal}{C\#} (K0613) \\
    \bottomrule
    \end{tabularx}%
    }\caption{
    We show six qualitative examples. The mention is indicated with purple and we show the predictions ($\mathbf{k = 1}$) of BLINK and GENRE. Green predictions mean correct, and red indicates wrong linking with respect to the ground truth. We also show the ESCO ID, indicating the differences in concepts. The results show successful linking of implicit mentions of skills. In example (5), we show how the linked results are still valid while being different concepts. However, evaluation does not count it as a correct hit.}
    \label{tab:quali}
\end{table*}

\section{Results}
\cref{tab:results} presents the results. Each model is trained for five seeds, and we report the average and standard deviation. We make use of a random and TF-IDF-based baseline. 

Firstly, we observe that the strict linking performance (i.e., Acc@1) is rather modest for both BLINK and GENRE. But most models outperform the baselines. Notably, the top-performing models in this context are the \bertl{} and \bartl{} models, which were further fine-tuned from Wikipedia EL with ESCO. As expected, scores improve considerably as we increase the value of $k$.  Secondly, for both BLINK and GENRE, model size seems not to have a substantial impact when trained only on ESCO. Specifically for BLINK, the performance remains consistent for Acc@1 and exhibits only a slight decline as we relax the number of candidates for performance evaluation. For GENRE, the observed trend remains largely unchanged, even with a larger $k$.

Remarkably, the zero-shot setup performance of both BLINK and GENRE, when trained on Wikipedia, surpasses that of models trained solely on ESCO. For Wikipedia-based evaluation, GENRE usually outperforms BLINK. We notice the opposite in this case. For BLINK, this improvement is approximately 11 accuracy points for $k = 1$. Meanwhile, for GENRE, we observe an increase of roughly 9 accuracy points when trained on both Wikipedia and ESCO. This trend persists for a larger $k$, reaching up to a 12.5 accuracy point improvement for BLINK and a 34 accuracy point improvement for GENRE in the case of Acc@32. Furthermore, we show that further fine-tuning the Wikipedia-trained models on ESCO contributes to an improved EL performance at $k = \{1, 16, 32\}$ for both models. We confirm our hypothesis that Wikipedia has concepts that are also in ESCO, this gives the model strong prior knowledge of skills.

For \texttt{UNK}-specific results, we refer to~\cref{app:unk}. Additionally, we show a direct comparison to previous work in~\cref{app:previous}.

\section{Discussion}\label{sec:discussion}

\paragraph{Qualitative Analysis.}
We manually inspected a subset of the predictions. We present qualitative examples in~\cref{tab:quali}. We found the following trends upon inspection: 
\begin{itemize}
    \itemsep0em
    \item The EL models exhibit success in linking implicit and explicit mentions to their respective taxonomy titles (e.g., \circled{1}, \circled{2}, \circled{4}, \circled{6}).
    \item In cases of hard skills (\circled{3}, \circled{6}), BLINK correctly matches ``MS office tools'' to ``using Microsoft Office'', which is not an exact match. Both models predict the explicit mention ``C\#'' correctly to the C\# taxonomy title.
    \item We found that the models predict paraphrased versions of skills that could also be considered correct (\circled{4}, \circled{5}), even being entirely different concepts (i.e., different ESCO IDs).
\end{itemize}

\paragraph{Evaluation Limitation.}
We qualitatively demonstrate the linking of skills that are implicit and/or valid.
Empirically, we observe that the strict linking of skills leads to an underestimation of model performance. We believe this limitation is rooted in evaluation. In train, dev, and test, there is only \emph{one} correct gold label. We reciprocate the findings by~\citet{li-etal-2020-efficient}, where they found that a large number of predictions are ``technically correct'' but limitations in Wikipedia-based evaluation falsely penalized their model (i.e., a more or less precise version of the same entity).
Especially \circled{5} in~\cref{tab:quali} shows this challenge for ESCO, we can consider multiple links to be correct for a mention given a particular context. This highlights the need for appropriate EL evaluation sets, not only for ESCO, but for EL in general.

\section{Conclusion}
We present entity linking in the job market domain, using two existing high-performing neural models. We demonstrate that the bi-encoder architecture of BLINK is more suited to the job market domain compared to the autoregressive GENRE model. While strict linking results favor BLINK over GENRE, if we relax the number of candidates, we observe that GENRE performs slightly better. From a qualitative perspective, the performance of strict linking results is modest due to limitations in the evaluation set, which considers only one skill correct per mention. However, upon examining the predictions, we identify valid links, suggesting the possibility of multiple correct links for a particular mention, highlighting the need for more comprehensive evaluation. We hope this work sparks interest in entity linking within the job market domain.

\section*{Limitations}
In the context of EL for ESCO, our approach has several limitations. Firstly, it only supports English, and might not generalize to other languages. However, several works are working on multilingual entity linking (e.g.,~\citealp{botha-etal-2020-entity, de-cao-etal-2022-multilingual}) and ESCO itself consists of 28 European languages. This work could be extended by supporting it for more languages.

Secondly, our EL model is trained on synthetic training data, which may not fully capture the intricacies and variations present in real-world documents. The use of synthetic data could limit its performance on actual, real JD texts. Nevertheless, we have human-annotated evaluation data.

Moreover, in our evaluation process, we use only one gold-standard ESCO title as the correct answer. This approach may not adequately represent a real-world scenario, where multiple ESCO titles could be correct as shown in~\cref{tab:quali}.

In~\cref{tab:results}, we show that providing in-domain data for continuous pre-training shows larger improvements for GENRE than for BLINK. We did not conduct a detailed analysis on the underlying reasons for these positive variations.

\section*{Acknowledgements}
We thank the MaiNLP and NLPnorth group for feedback on an earlier version of this paper, and the reviewers for their insightful comments. This research is supported by the Independent Research Fund Denmark (DFF) grant 9131-00019B and in parts by ERC Consolidator Grant DIALECT 101043235. 

\bibliography{anthology,custom}

\appendix

\begin{algorithm*}[t]

\KwData{%
  $sentence$: The input sentence \\
  $target\_subspan$: The target subspan \\
  $threshold$: The Levenshtein distance similarity threshold \\
}

\KwResult{%
  $most\_similar\_ngram$: The most similar n-gram \\
}

\BlankLine

$all\_ngrams \leftarrow$ GenerateAllNgrams($sentence$)

$filtered\_ngrams \leftarrow$ FilterNgrams($all\_ngrams$, $target\_subspan$, $threshold$)

$most\_similar\_ngram \leftarrow$ None \\
$max\_similarity \leftarrow 0$

\For{$ngram$ in $filtered\_ngrams$}{
    $subspan\_embedding \leftarrow$ EncodeWithSBERT($target\_subspan$) \\
    $ngram\_embedding \leftarrow$ EncodeWithSBERT($ngram$)
    
    $similarity \leftarrow$ CosineSimilarity($subspan\_embedding$, $ngram\_embedding$)
    
    \If{$similarity > max\_similarity$ \textbf{and} $similarity > 0.5$}{
        $max\_similarity \leftarrow similarity$ \\
        $most\_similar\_ngram \leftarrow ngram$
    }
    \Else{$most\_similar\_ngram$ = UNK}
}

\Return $most\_similar\_ngram$

\caption{Find the most similar n-gram to a target subspan}
\label{alg:find-similar-ngram}
\end{algorithm*}

\section{Data Preprocessing}\label{app:preprocess}

We outline the preprocessing steps for the training set. In~\citet{decorte2023extreme}, there are sentence--ESCO skill title pairs. The data is synthetically generated by GPT-3.5. Where for each ESCO skill title a set of 10 sentences is generated. A crucial limitation for entity linkers is that the generated sentence does not have the ESCO skill title as an exact match in the sentence, but at most slightly paraphrased. To find the most similar subspan in the sentence to the target skill, we have to apply some heuristics. In Algorithm~\ref{alg:find-similar-ngram}, we denote our algorithm to find the most similar subspan. Our method is a brute force approach, where we create all possible n-grams until the maximum length of the sentence, and compare the target subspan against each n-gram. Based on Levenshtein distance, we filter the results, where we only take the top 80\% n-grams. Then, we encode both target subspan and n-gram with SentenceBERT~\cite{reimers-gurevych-2019-sentence}, the similarity is based on cosine similarity. If the similarity does not exceed 0.5, the candidate subspan is \texttt{UNK} and the ESCO title will also be \texttt{UNK}, otherwise, we take the most similar n-gram. Empirically, we found that these thresholds worked best. Note that this method is not error-prone, but allows us to generate implicit and negative examples to train entity linkers. We show two qualitative examples in~\cref{lst:example_train} and discuss the quality in~\cref{app:examples}.

\begin{table*}[ht]
    \centering
    \caption{\textbf{\texttt{UNK} Linking Results.} We show the results of BLINK and GENRE predicting \texttt{UNK}. We use the best-performing models, based on~\cref{tab:results}.}
    \resizebox{.9\textwidth}{!}{%
    \begin{tabular}{lrrrrrrr}
    \toprule
         & \textbf{Train Source} & \textbf{Acc@1}  & \textbf{Acc@4} & \textbf{Acc@8} & \textbf{Acc@16} & \textbf{Acc@32} \\
    \midrule
BLINK (\texttt{bert-large}) \texttt{UNK} & Wiki + ESCO & \std{1.38}{0.12}  & \std{3.32}{0.22} & \std{4.67}{0.33} & \std{7.68}{0.42} & \std{10.70}{0.58} \\\midrule
GENRE (\texttt{bart-large}) \texttt{UNK} & Wiki + ESCO & \std{1.65}{0.20}  & \std{4.99}{0.50} & \std{9.23}{0.58} & \std{16.01}{0.48} & \std{24.70}{2.52} \\    \bottomrule
    \end{tabular}%
    }
    \label{tab:unk}
\end{table*}

\begin{figure*}[t]
\begin{minipage}{\textwidth}
\begin{minted}[frame=single,
               framesep=3mm,
               linenos=true,
               xleftmargin=15pt,
               tabsize=2]{json}
{
    "context_left": "we're looking for someone who is passionate 
    about", 
    "context_right": "and eager to share their knowledge with 
    others.", 
    "mention": "young horse training", 
    label_title": "young horses training", 
    "label": "Principles & techniques of educating young horses 
    important simple body control exercises.", 
    "label_id": 2198
}
{
    "context_left": "Hands-on experience with", 
    "context_right": "is a must-have qualification for this 
    job.", 
    "mention": "various hand-operated printing devices", 
    "label_title": "types of hand-operated printing devices", 
    "label": "Process of creating various types hand-operated 
    printing devices, such as stamps, seals, embossing labels or 
    inked pads and their applications.", 
    "label_id": 10972
}
\end{minted}
\caption{\textbf{Two Training Examples.} The training examples are in the format for BLINK, there is the left context, right context, and the mention. The label title is the ESCO skill, and the label is the description of the label title. The label ID is the ID that refers to the label title.}
\label{lst:example_train}
\end{minipage}
\end{figure*}

\begin{figure*}[t]
\begin{minipage}{\textwidth}
\begin{minted}[frame=single,
               framesep=3mm,
               linenos=true,
               xleftmargin=15pt,
               tabsize=2]{json}
{
    "context_left": "You must have an", 
    "context_right": "with a high-quality mindset.", 
    "mention": "analytical proactive and structured workstyle", 
    "label_title": "work in an organised manner", 
    "label": "Stay focused on the project at hand, at any time. 
    Organise, manage time, plan, schedule and meet deadlines.", 
    "label_id": 3884
}
\end{minted}
\end{minipage}
\caption{\textbf{One Evaluation Example.} The evaluation example is in the format for BLINK, there is the left context, right context, and the mention. The label title is the ESCO skill, and the label is the description of the label title. The label ID is the ID that refers to the label title.}
\label{lst:example_dev}
\end{figure*}

\section{Data Examples}\label{app:examples}
We show a couple of data examples from the training (\cref{lst:example_train}) and development set (\cref{lst:example_dev}). In the training examples, we show an example with a mention that is the same as the original ESCO title (``young horse training''). In addition, we have an example where there is an ``implicit'' mention (i.e., the mention does not exactly match with the label title). This shows that our algorithm works to an extent. For the development example, this is another implicit mention. However, these samples are human annotated. 
There are also quite some \texttt{UNKs} given the training data. We show that this is helping the model predict \texttt{UNK}.

\section{Implementation Details}\label{app:hyperparams}
For training both BLINK\footnote{\url{https://github.com/facebookresearch/BLINK}} and GENRE,\footnote{\url{https://github.com/facebookresearch/genre}} we use their respective repositories. All models are trained for 10 epochs, for a batch size of 32 for training and 8 for evaluation. For both BLINK and GENRE we use 5\% warmup. For the base models we use learning rate $2 \times 10^{-5}$ and for the large models we use $2 \times 10^{-6}$. The maximum context and candidate length is 128 for both models. Each model is trained on an NVIDIA A100 GPU with 40GBs of VRAM and an AMD Epyc 7662 CPU. The seed numbers the models are initialized with are 276800, 381552, 497646, 624189, 884832. We run all models with the maximum number of epochs (10) and select the best-performing one based on validation set performance for accuracy@1.

\section{\texttt{UNK} Evaluation}\label{app:unk}
In~\cref{tab:unk}, we show the performance of both BLINK and GENRE on the \texttt{UNK} label. We use the best-performing models based on~\cref{tab:results}. Generally, we observe that GENRE is better in predicting \texttt{UNKs} than BLINK. However, the exact linking results (i.e., Acc@1) are low. This can potentially be alleviated by actively training for predicting \texttt{UNKs}~\cite{zhu-etal-2023-learn}.

\section{Comparison To Previous Work}\label{app:previous}
We argue that an entity linking approach to match skill spans to ESCO taxonomy codes is the correct direction as it could provide more transparency in the linked span in the sentence. Consequentially, this is a more challenging setup. In~\cref{tab:comparison}, we provide a direct comparison to previous work from~\citet{decorte2023extreme} and~\citet{clavie2023large}, where they link sentences with skills directly. For context, we are not using re-rankers as in the previously mentioned works.

\begin{table*}[t]
    \centering
    \begin{tabular}{llr}
    \toprule
    \textbf{Approach}     & \textbf{Setup}      & \textbf{MRR} \\
    \midrule
    \citet{decorte2023extreme}  & \texttt{SentenceBERT, sentence-level, re-ranking}             &  \std{47.8}{0.0}\\
    \citet{clavie2023large}     & \texttt{GPT4, sentence-level, re-ranking}                     &  \std{51.6}{0.0}\\
    \midrule
    \textbf{This work}          & \texttt{BLINK, mention-level, no re-ranker}                   &  \std{28.8}{0.1}\\
    \textbf{This work}          & \texttt{GENRE, mention-level, no re-ranker}                   &  \std{17.5}{0.2}\\
    \bottomrule
    \end{tabular}
    \caption{We show a comparison to previous work, in a more challenging setup. We measure the performance in mean reciprocal rank (MRR). Note that previous work separates the splits in the ESCO matching dataset by~\citet{decorte2023extreme}, we average them here. We highlight the differences in setup, which indicates the unfair comparison. We show the results of the best-performing models (i.e., BLINK/GENRE \texttt{large} with Wikipedia and ESCO as training data).}
    \label{tab:comparison}
\end{table*}

\end{document}